\documentclass[conference]{IEEEtran}

\usepackage{amsmath,amssymb,amsfonts,amsthm}
\usepackage{algorithmic}
\usepackage{array}
\usepackage{textcomp}
\usepackage{stfloats}
\usepackage{url}
\usepackage{verbatim}
\usepackage{subcaption}

\usepackage{graphicx}

\usepackage{diagbox}
\usepackage{siunitx} 
\ifCLASSOPTIONcompsoc
    \usepackage[caption=false, font=normalsize, labelfont=sf, textfont=sf]{subfig}
\else
\usepackage[caption=false, font=footnotesize]{subfig}
\fi

\usepackage{cite}
\usepackage{multirow}
\usepackage{multicol}

\usepackage[utf8]{inputenc}
\usepackage{enumitem}

 \usepackage{fancyhdr}
\fancypagestyle{mystyle}{
    \chead{\large The 2024 IEEE International Conference on Big Data (IEEE BigData 2024)}}

\usepackage[linesnumbered,ruled,vlined]{algorithm2e}

\SetKwComment{Comment}{$\triangleright$\ }{}

\usepackage{nomencl}
\makenomenclature
\renewcommand{\nompreamble}{The next list describes several symbols that will be later used within the body of the Article}
\makeatletter
\newcounter{subparagraph}[paragraph]
\renewcommand\thesubparagraph{%
  \theparagraph.\@arabic\c@subparagraph}
\newcommand\subparagraph{%
  \@startsection{subparagraph}    
    {5}                              
    {\parindent}                     
    {2ex \@plus 1ex \@minus .6ex} 
    {0em}                           
    {\normalfont\normalsize\bfseries}}
\usepackage{ragged2e}  
\usepackage{blindtext}
\usepackage[T1]{fontenc}
\usepackage[utf8]{inputenc}
\usepackage[spaces,hyphens]{xurl}
\newcommand*{\Resize}[2]{\resizebox{#1}{!}{$#2$}}%

\newcommand{\fref}[1]{Fig.~\ref{#1}}

\newcommand\HUGE{\fontsize{21.4 }{25}\selectfont}

\begin{document}

\title{\HUGE  FisherMask: Enhancing Neural Network Labeling Efficiency in Image Classification Using Fisher Information }
\author{\IEEEauthorblockN{
Shreen Gul\IEEEauthorrefmark{2}, Mohamed Elmahallawy\IEEEauthorrefmark{3}, Sanjay Madria\IEEEauthorrefmark{2}, Ardhendu Tripathy\IEEEauthorrefmark{2}}  
      \IEEEauthorblockA{%
 \IEEEauthorrefmark{2}Computer Science Department, Missouri University of Science and Technology, Rolla, MO 65401, USA}
       \IEEEauthorblockA{%
 \IEEEauthorrefmark{3} {School of Engineering \& Applied Sciences, Washington State University, Richland, WA 99354, USA}}
Emails:  sgchr@mst.edu, mohamed.elmahallawy@wsu.edu, madrias@mst.edu, astripathy@mst.edu}

\maketitle
\thispagestyle{mystyle}

\begin{abstract}
Deep learning (DL) models are popular across various domains due to their remarkable performance and efficiency. However, their effectiveness relies heavily on large amounts of labeled data, which are often time-consuming and labor-intensive to generate manually. To overcome this challenge, it is essential to develop strategies that reduce reliance on extensive labeled data while preserving model performance. In this paper, we propose FisherMask, a Fisher information-based active learning (AL) approach that identifies key network parameters by masking them based on their Fisher information values. FisherMask enhances batch AL by using Fisher information to select the most critical parameters, allowing the identification of the most impactful samples during AL training. Moreover, Fisher information possesses favorable statistical properties, offering valuable insights into model behavior and providing a better understanding of the performance characteristics within the AL pipeline. Our extensive experiments demonstrate that FisherMask significantly outperforms state-of-the-art methods on diverse datasets, including CIFAR-10 and FashionMNIST, especially under imbalanced settings. These improvements lead to substantial gains in labeling efficiency. Hence serving as an effective tool to measure the sensitivity of model parameters to data samples. Our code is available on \url{https://github.com/sgchr273/FisherMask}.


\end{abstract}


\section{Introduction}
Deep learning (DL) networks are increasingly integrated into numerous fields due to their remarkable performance and accuracy. However, their efficacy heavily relies on labeled/annotated data. Manual annotation of data is often costly, prompting a growing demand for techniques capable of achieving high performance with limited labeled data \cite{yuan2023active}. Active learning (AL) emerges as one such strategy, exploiting informative samples to train models and thereby diminishing the necessity for additional data, thus mitigating the demand for more annotated data \cite{nemeth2024compute}. Scenarios such as medical imaging, speech recognition, and anomaly detection tasks can greatly benefit from AL \cite{10386969,chen2022supervised,tang2020deep}.

AL approaches, such as those proposed by \cite{ash2021gone, hino2023active, li2024unlabeled}, have been developed to gauge the informativeness of data samples and select them for model training. These methods utilize information-theoretic measures like Fisher information, entropy, and Kullback-Leibler divergence to assess the significance of samples within a dataset \cite{hino2023active}. This work will center on Fisher information, which measures how much information an observable random variable reveals about an unknown parameter in a distribution. Fisher information is an effective method due to its independence from ground truth values and its always semi-definite nature. These properties make it advantageous in various applications, including optimization, control theory, and machine learning \cite{ly2017tutorial}.

Different query strategies use this Fisher information measure in various ways: some focus on selecting uncertain samples, while others prioritize diversity in sample selection. Some approaches approximate Fisher information values through trace operations. For instance, the method introduced in \cite{ash2021gone}, known as BAIT (\underline{\bf B}atch \underline{\bf A}ctive learning via \underline{\bf I}nformation ma\underline{\bf T}rices), employs a Fisher-based greedy approach. This method selects samples by minimizing an objective function that incorporates approximations of Fisher information matrices and their inverses. Another work presented in \cite{sung2021training} proposes training the network by updating only a subset of parameters rather than all of them. They reported an approximation of parameter importance based on the average squared gradients of the model’s output. This approximation helps quantify the significance of each parameter. In another work, the authors of \cite{zeiler2014visualizing} noted that higher layers of deep networks are better at generating discriminative features compared to lower layers. Furthermore, \cite{chen2022layer} showed that deeper layers capture more complex aspects of the target function. These insights motivate our proposed approach, which leverages the discriminative power of upper layers to enhance feature learning from the dataset. Incorporating these layers into our process could potentially lead to better and more informative samples.

\begin{figure*}[!t]
    \centering
    \includegraphics[width= 0.8\textwidth]{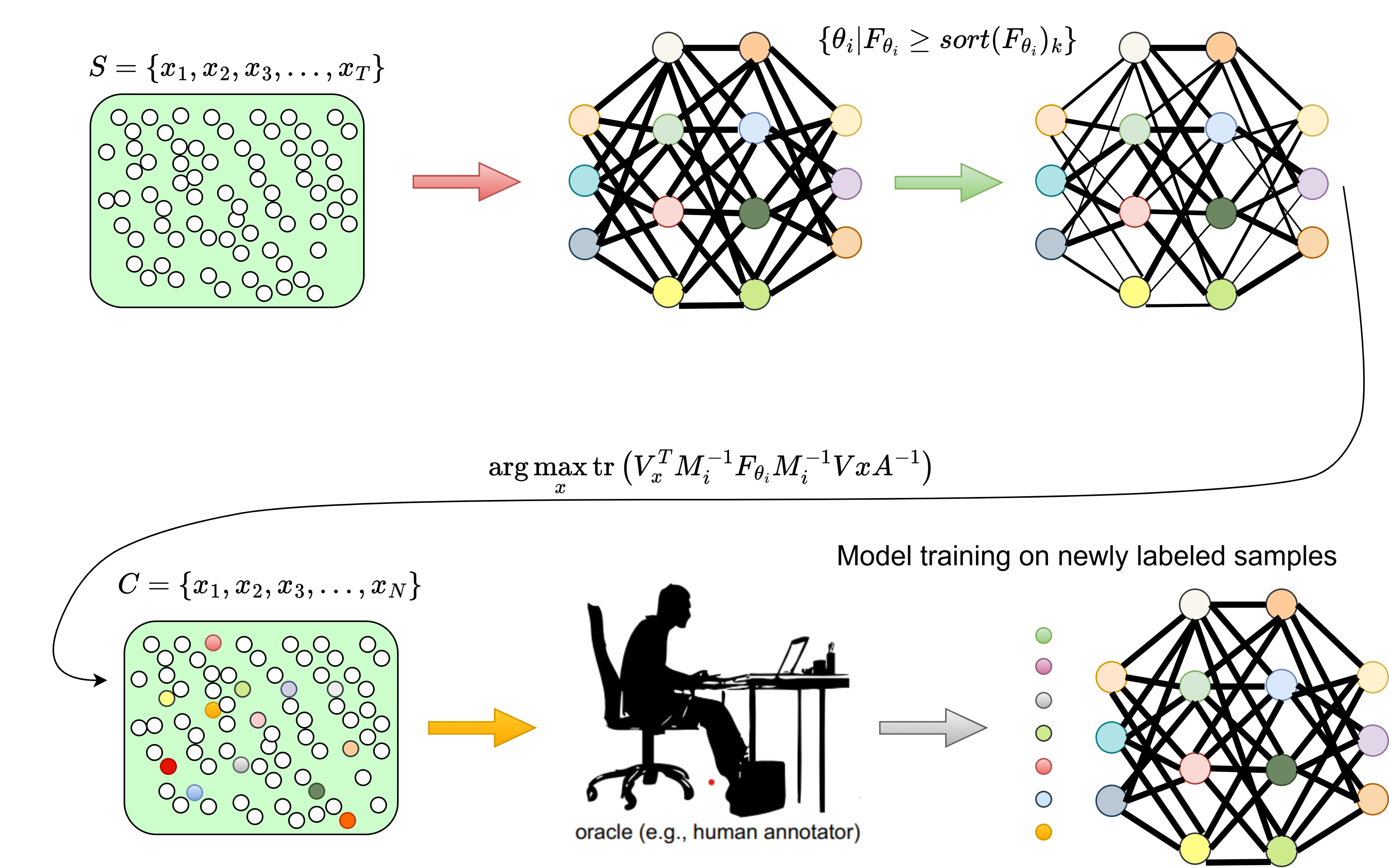}
\caption{Illustration of important weights sampling. Hollow circles represent the set of unlabeled samples \( \mathcal{S} \) fed into the neural network. Colored arrows depict the process of identifying important weights while pruning the remaining ones. Based on the selected weights, a subset of unlabeled instances \( \mathcal{C} \) (colored circles) is chosen for labeling. This subset is then sent to an oracle for labeling, after which the model will be trained on this newly labeled data, shown in the lower-left portion of the figure, completing one AL round.}
\label{fig:1}
\end{figure*}

{\bf Contributions.} Motivated by the work of \cite{sung2021training}, we develop a method called FisherMask for constructing a sparse network mask. FisherMask aims to leverage Fisher information to capture crucial details about unlabeled data samples. As illustrated in \fref{fig:1}, we compute the Fisher information matrix for the entire network and use it to create this mask, which is why we refer to it as FisherMask. This mask is formed by selecting $k$ weights with the highest Fisher information values. To speed up computations, we approximate the updates to the Fisher information matrices and their inverses using the Woodbury identity and trace rotation techniques, similar to those used in BAIT \cite{ash2021gone}. FisherMask specifically utilizes the information from the network's middle layers to identify influential samples. To sum up, our contributions  are:
\begin{enumerate}[label=(\arabic*),leftmargin=*]
    \item We propose FisherMask, a novel method for constructing a sparse network mask based on Fisher information. This method leverages {\em important weights} to capture critical details about unlabeled data samples by selecting the $k$ weights with the highest Fisher information values, specifically for large datasets with limited labels.

    \item To enhance computational efficiency, we approximate updates to the Fisher information matrices and their inverses using the Woodbury identity and trace rotation techniques. This approach leverages information from the network's middle layers to effectively identify influential samples.

    \item Our performance evaluations on a range of diverse and publicly available datasets highlight the effectiveness and model-agnostic nature of FisherMask. Additionally, the results show that FisherMask achieves performance that is comparable to or exceeds that of existing methods, particularly in scenarios with imbalanced datasets.
    
\end{enumerate}
\section{Related Work}
AL encompasses a range of techniques aimed at making the training of machine learning (ML) models more efficient by strategically selecting which data points to label. These techniques generally fall into two primary categories: {\em uncertainty sampling} and {\em diversity sampling}. Below is a detailed overview of each approach.

{\bf Uncertainty-based approaches.} They seek to reduce labeling efforts by focusing on samples where the model is most uncertain. Key techniques in this area include entropy sampling, margin sampling, and mutual information \cite{wang2014new}. However, each of these methods has its limitations. For example, entropy sampling can overlook the interrelationships between samples, leading to the selection of redundant data. Similarly, mutual information, while theoretically informative, often involves high computational complexity, making it less practical for high-dimensional datasets \cite{sourati2017asymptotic}.

{\bf Diversity-based approaches} They focus on selecting samples that effectively represent the overall distribution/diversity of the entire dataset.  Some popular approaches in this category include k-means sampling \cite{zhdanov2019diverse}, k-means++ \cite{ash2019deep}, and k-center greedy (also known as coreset) \cite{sener2017active}. However, each of these approaches comes with its own limitations. For instance, the k-center greedy method, while useful for identifying diverse samples, often faces significant computational challenges  and ends up taking a lot more computational time than any other AL strategy. The process involves constructing a distance matrix for each unlabeled sample, which can be 
resource-intensive \cite{shui2020deep}.

Some studies {\em combine both model uncertainty and dataset diversity} to select the most informative samples for AL. For example, \cite{ash2019deep} introduces a hybrid approach called BADGE (\underline{\bf B}atch \underline{\bf A}ctive learning by \underline{\bf D}iverse \underline{\bf G}radient \underline{\bf E}mbeddings). This method evaluates uncertainty by measuring the gradient length concerning the network's last-layer parameters while ensuring diversity through k-means++ clustering. BADGE effectively leverages data embeddings, which is advantageous when feature learning is a key benefit of deep neural networks \cite{kirsch2022unifying}.



\section{Preliminaries}
\subsection{AL via information matrices}
The process selects the next optimal sample to include in the labeled set by optimizing the objective function such as the one 
in \cite{ash2021gone}. Specifically, this function aims to maximize the  potential information by  informativeness of the newly labeled sample, considering both the current model uncertainty and the diversity of the data. For reference, the objective function can be given by:
\begin{equation}
\label{eq:1}
\mathcal{O}(\mathbf{x}) = \arg\max_{x} \operatorname{tr} \left( V_x^T M_i^{-1} F^L M_i^{-1} V_x A^{-1} \right),
\end{equation}
where \( V_x \) is the matrix of gradients of the model's predictions with respect to the parameters, and \( M_i \) is the Fisher information matrix of the labeled samples, given by:
\begin{equation}
       M_i = \lambda F + \frac{1}{|C|} \sum_{x \in C} F(x; \theta^L),
\end{equation}
where \( C \) is the set of selected samples, and \( \lambda \) is a regularization parameter. \(A\) is an adjustment to the equation and reflects not only the general Fisher information but also how the sample-specific gradients impact the model's uncertainty or information density in the parameter space, which can be given as:
\begin{equation}
A = F + V_x^T M_i^{-1} V_x,
\end{equation}
where \( F \) is the Fisher information matrix and \( V_x^T M_i^{-1} V_x \) represents the contribution of the gradients. \( F^L \) denotes the Fisher Information matrix of the last layer of the network, which can be expressed as:
\begin{equation}
     F^L = \mathbb{E}_{y \sim p( x \mid \theta^L)} \left[ \nabla^2 l(x, y; \theta^L) \right]
\end{equation}
Here \( \theta^L \) denotes the weights of the last layer of the network and \( l(x, y; \theta^L) \) is the loss function.


\subsection{Entropy sampling}
In AL, entropy sampling is used to select the most uncertain data points for labeling. The uncertainty is quantified by the entropy of the model's predicted probability distribution. The data points with the highest entropy are considered the most informative and thus prioritized for labeling. The entropy \( H \) of a data point is calculated using the formula:
\begin{equation}
\label{eq:entropy_sampling}
H = -\sum_{i=1}^{N} p_i \log_2(p_i),
\end{equation}
where \( N \) is the number of classes in the dataset, and \( p_i \) is the probability of the data sample belonging to class \( i \). Higher entropy values indicate greater uncertainty in the model’s predictions, making such samples more valuable for improving the model's performance through AL.

\subsection{Margin Sampling} 
Margin Sampling is a method used to select data samples where the model's prediction is the least confident. This is achieved by focusing on the margin between the probabilities of the top two predicted classes.
\begin{equation}
\label{eq:margin_sampling}
M(x) = |p_{\text{max}}(x) - p_{\text{second}}(x)|
\end{equation}
where $M(x)$ denotes the margin of the model's prediction for sample $x$, which quantify how close the model's prediction is to being uncertain, with a smaller margin indicating higher uncertainty. $p_{\text{max}}(x)$ is the probability score of the top predicted class for sample $x$, and $p_{\text{second}}(x)$ is the probability score of the second most probable class for sample $x$.

\subsection{$k$-center greedy}
In this algorithm, $b$ points are selected from a set $S$ as center points to minimize the maximum distance between any data point $x_i$ and its nearest center $x_j$ \cite{settles2009active}. Mathematically, the problem can be formulated as:
\begin{equation}
\min_{S^1: |S^1| \leq b} \max_{i} \left( \min_{j} \Delta(x_i, x_j) \right)
\end{equation}
where $S^1$ is the set of selected center points with at most $b$ elements, $x_i$ represents a data point in the set, $x_j$ represents a center point from the selected set $S^1$, and $\Delta(x_i, x_j)$ denotes the distance between data point $x_i$ and center point $x_j$.




\section{Methodology}
\subsection{Problem Statement}

In this work, we aim to optimize the training process of an ML  model $f$ using a dataset \(S = \{x_1, x_2,... x_T\}\). The objectives are twofold:
\begin{enumerate}
    \item {\em Identify the most important network parameters:} We calculate the Fisher information matrix (FIM) for the model's parameters \(\theta\). The FIM is defined as:
    \begin{equation}\label{eq:fisher_info}
    \Resize{7.3cm}{\text{FIM}_{i} = \frac{1}{T} \sum_{i=1}^{T}\mathbb{E}_{y \sim p_{\theta}(y | x_{i})}(\nabla_{\theta}\log p_{\theta}(y|x_{i}))^2    ,}
    \end{equation}

    where \(\text{FIM}_{i}\) measures the sensitivity of the log-likelihood with respect to the model parameters \(\theta_i\). By analyzing the FIM, we identify which parameters have the greatest impact on the model’s performance and learning dynamics.

    \item {\em Determine the Most Influential Samples:} We evaluate which samples in \(S\) are most influential for the model. This involves assessing the contribution of each sample to the overall learning process, which can be guided by metrics such as gradient norms, influence functions, or other techniques that measure the impact of individual samples on the model's parameters.
\end{enumerate}


\subsection{Notations}
We focus on the standard batch AL scenario involving the instance space \textit{X}, the label space \textit{Y}, and the distribution \(D_{Y|X}(x)\) of the label space given an input \(x\). We have access to a set of unlabeled data \(S = \{x_1, x_2, \ldots, x_T\}\), from which we can selectively request a batch of \(N\) data points for labeling. In the \(p\)-th AL cycle, we select a collection \(C = \{x^{(p)}_k\}_{k=1}^N\) of \(N\) samples and request their labels \(y^{(p)} \sim D_{Y|X}(x_{k}^{(p)})\) from the oracle. Our primary objective is to minimize the following loss function as:
\begin{equation}
    L_{S}(\theta) = \mathbb{E}_{x \sim S, y \sim D_{Y|X}(x)} \left[ l(x, y; \theta^*) \right],
\end{equation}
where \(S\) is the set of unlabeled data, \(\theta^*\) denotes the learned model parameters, and \(l(x, y; \theta^*)\) is the loss function associated with the model's prediction for input \(x\) and true label \(y\). The goal is to achieve this with the fewest possible data points. In this context, we treat the unlabeled data \(S\) as representative of the entire distribution and utilize the FIM to perform AL on the given unlabeled set \(S\).

From \eqref{eq:fisher_info}, it is evident that a particular element in the FIM represents the average of the squared gradients of the network's predictions \(y\) concerning its parameters \(\theta\). Specifically, if a parameter significantly influences the model's output, its corresponding element in the FIM will be large. Therefore, Fisher information can be effectively used to measure the importance of the network's parameters.

Motivating by \cite{chaudhuri2015convergence}, we pose our objective function as follows: 
\begin{equation}\label{eq:obj}
    \tilde{x} = \underset{x \in S}{\arg\min \operatorname{tr} }\left((M_i + F(x; \theta_{r}))^{-1}F_{\theta_{r}}\right)
\end{equation}
where \(\theta_{r}\) represents sparsely selected weights in $r_{th}$ AL round. $F_{\theta_{r}}$ is Fisher information of unlabeled samples, which can be expressed as \(F_{\theta_{r}} = \frac{1}{T} \sum_{i=1}^{T}\mathbb{E}_{y \sim p_{\theta}(y | x_{i})}(\nabla_{\theta_{r}}\log p_{\theta}(y|x_{i}))^2\) with $M_{i}$
represents the \(i\)-th labeled sample to be included in the collection \(C\) and is given by \(M_i = \lambda F + \frac{1}{|C|}\sum_{x \in C}F(x;\theta_r)\).

We employ the Woodbury identity and trace rotation for inverse updates, a technique analogous to that used in \cite{ash2021gone}, to approximate the expression in \eqref{eq:obj}. The algebraic expressions for these updates can be provided as
\begin{align}
\tilde{x} &=\underset{x}{\arg\min \operatorname{tr}} \big( ( M_i + V_{x}V_{x}^T )^{-1} F_{\theta_{r}} \big) \\ \nonumber 
&= \underset{x}{\arg\min \operatorname{tr}} \big( ( M_i^{-1} - M_{i}^{-1}V_{x}A^{-1}V^{T}M_{i}^{-1} )F_{\theta_{r}} \big)\\
&= \underset{x}{\arg\min \operatorname{tr}} \big( M_{i}^{-1}F_{\theta_{r}}\big) - \text{tr} \big( M_{i}^{-1}V_{x}A^{-1}V_{x}^{T}M_{i}^{-1}F_{\theta_{r}}  \big) \nonumber\\
&=\underset{x}{\arg\min \operatorname{tr}} \big( M_{i}^{-1}F_{\theta_{r}}\big) - \text{tr} \big(V_{x}^{T}M_{i}^{-1}F_{\theta_{r}}M_{i}^{-1} V_{x} A^{-1} \big) \nonumber
\end{align}
After applying the linear algebra techniques, an approximate solution to the optimization problem defined in \eqref{eq:obj}, through which we select the next best sample, is given by:
\begin{equation}\label{eq:sol}
     \arg\max_{x} \operatorname{tr} \left( V^T_x M_i^{-1}F_{\theta}M_i^{-1}V_{x}A^{-1} \right)
\end{equation}
\begin{algorithm}[!t]
\caption{FisherMask Training Process}\label{alg:reward}
    \begin{algorithmic}[1]
    \renewcommand{\algorithmicrequire}{\textbf{Input}}  
    \renewcommand{\algorithmicensure}{\textbf{Output}}
    
    \REQUIRE {Model $f(x;\theta)$, pool of unlabeled examples S, AL rounds R}, sparsity parameter $k$
    \ENSURE {Learned model $\theta_R$} 
    \STATE Initialize set $C$ of points by selecting $N_o$ labeled samples from $S$ and fit the model on $C$: $\theta_{initial} = \arg\min_{\theta} \mathbb{E}_{S}[l(x,y;\theta)] $
    
    \FOR{r = 1, 2, ..., R}
        \STATE Calculate $F_{\theta_{r}} = \frac{1}{|S|}\sum_{x \in S}F(x;\theta_r)$
        \STATE Filter $F_{\theta_{r}}$ by $\{\theta | F_{\theta_{r}} \geq  sort(F_{\theta_{r}})_{k}\}$
        \STATE Initialize $ M_o = \lambda F + \frac{1}{|C|}\sum_{x \in C}F(x;\theta_r)$
        \FOR{n = 1, 2, 3, ..., N }
            \STATE $\tilde{x} = \underset{x \in S}{\arg\min \operatorname{tr} }\left((M_i + F(x; \theta_{r}))^{-1}F_{\theta_{r}}\right)$
            \STATE $M_{i+1} \leftarrow M_{i} + F(\tilde x;\theta_{r}), C \leftarrow \tilde x$
        \ENDFOR
        \STATE Train model on $C: \theta_{r} = \arg\min_{\theta} \mathbb{E}_{S}[l(x,y;\theta)] $
    \ENDFOR
\end{algorithmic}
\end{algorithm}
where $F_{\theta}$ represents the Fisher information of crucial parameters $\{\theta | F_{\theta} \geq  sort(F_{\theta})_{k}\}$ with $k$ signifies the level of sparsity for the selection of important weights. We tested multiple sparsity levels for constructing the FisherMask, including 0.01, 0.005, 0.002, and 0.001. Our experiments demonstrated that a sparsity level of 0.002 was optimal, yielding the best model performance. These sparsity levels were chosen relative to the total parameter count in our model architecture, specifically 11 million parameters in the case of ResNet-18. To derive the solution to Eqn. \eqref{eq:obj},  we employed the substitution $F(x;\theta_r) = V_xV_x^{T}$, which reduces the need to store all updates of $F(x;\theta)$ and improves  the computation cost. The invertible matrix $A = F + V_x^TM_{i}^{-1}V_x$ is of dimension $A \in \mathbb{R}^{n \times n}$. Here, $V_{x}$ is a matrix of size $\mathbb{R}^{mn \times n}$ containing gradients with each column scaled by the square root of the corresponding prediction. 
\begin{figure}[!!t]
\centering
\includegraphics[width=0.49\textwidth]{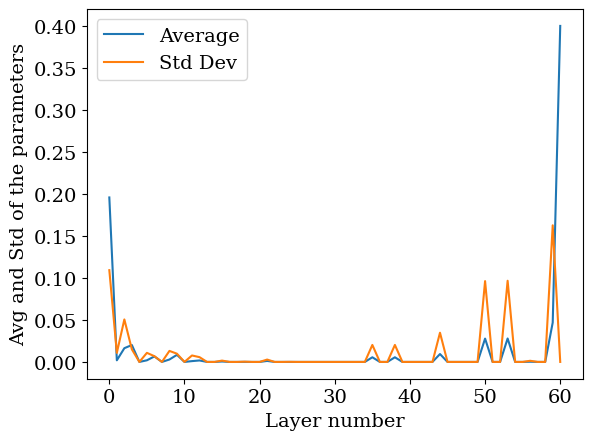}
\caption{Profile of important weights across Resnet-18}
\label{fig:2}
\end{figure}

In \fref{fig:2}, we illustrate the proportion of parameters deemed significant across the 61 layers of the ResNet18 model. The analysis reveals that weights from the initial 10 layers are consistently identified as significant. However, between layers 15 and 35, fewer parameters are deemed important by the algorithm. Notably, there are spikes in selection frequency in the later stages of the model, particularly from layer 35 onwards. The substantial spike at the 61st layer indicates that a significant fraction of weights from the final layer have been chosen for constructing the FisherMask.


Our approach differs from BAIT \cite{ash2021gone} in how we utilize Fisher information matrices to choose the optimal data points. Unlike BAIT, which relies solely on Fisher information matrices from the last layer of the network, we consider {\em weights across intermediate layers of the network}, as illustrated in \fref{fig:2}. The pseudocode for our proposed strategy is provided in Algorithm~\ref{alg:reward}. The Algorithm involves utilizing a classifier \( f \) along with an unlabeled set \( S \). Initially, the model is trained on randomly selected samples \( C \) as indicated in line 1. Subsequently, the AL process begins, where each cycle includes the calculation of Fisher Information values for both the chosen samples \( M \) and the remaining unlabeled pool \( F \), as outlined in lines 3 and 4. These computed values \( M \) and \( F \) are then used in Equation~\ref{eq:obj} to determine the optimal samples.

\begin{figure}
\centering
\includegraphics[width=0.65\textwidth]{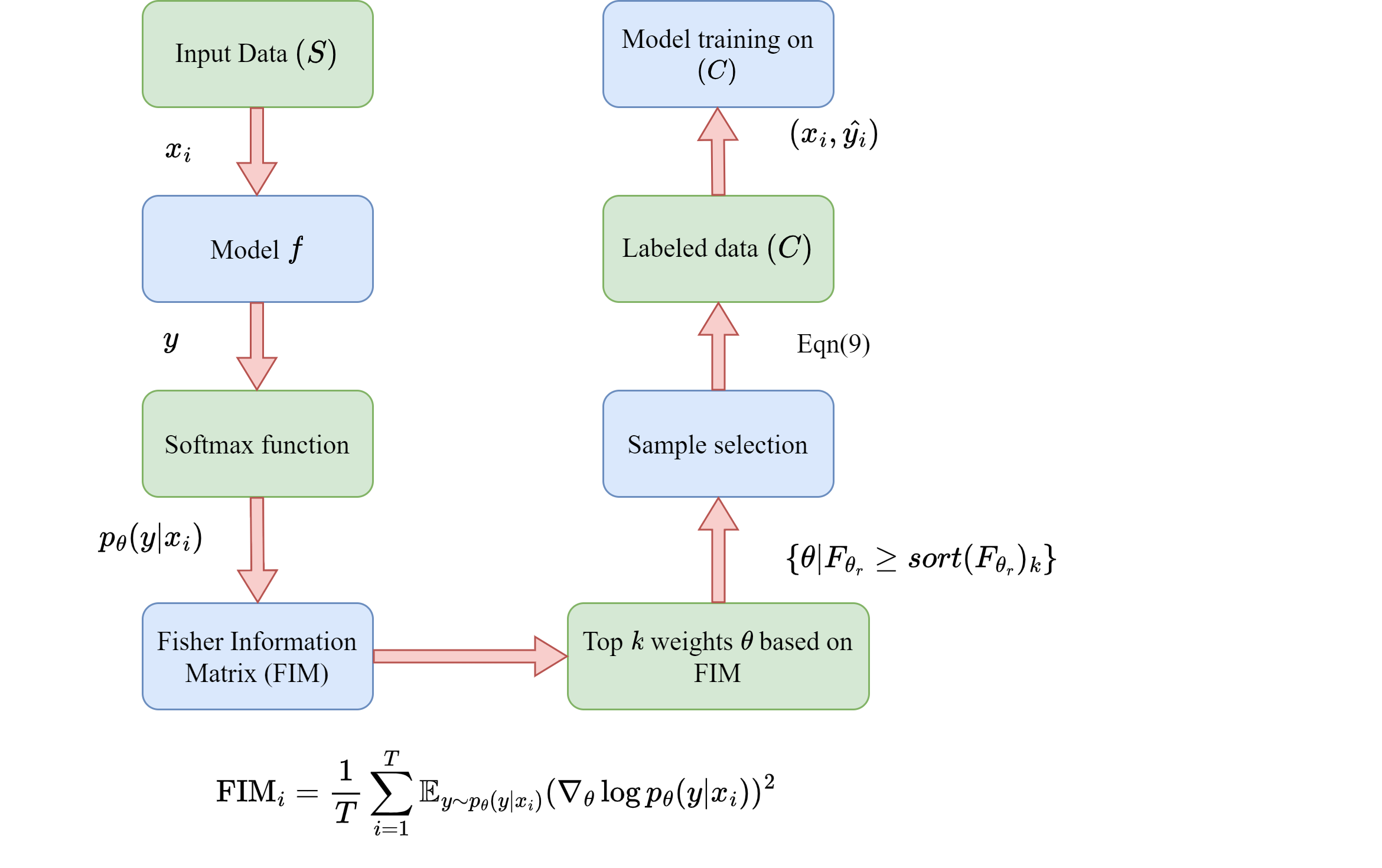}
\caption{Overview of FisherMask's framework.}
\label{fig:3}
\end{figure}

In \fref{fig:3}, we present the overall methodology for selecting a single data sample. A data point \( x_i \) from the sample space \( S \) is fed to the model to obtain a probability vector via the softmax layer. The FIM is calculated from this probability vector, and the top \( k \) parameters with the largest Fisher information values are selected. Using the objective function (Equation~\ref{eq:sol}), the next most influential sample is chosen to be added to the labeled dataset \( C \). The model is then trained on the set of these queried samples \((x_i, y_i)\). This process of selecting a batch of queried points is repeated a fixed number of times, and the cycle continues until the stopping criteria, such as the label budget, are met.



\section{Results and Discussion}

{\bf Datasets.} We utilize two datasets in our experiments: CIFAR-10 \cite{CIFAR-10} and FashionMNIST \cite{xiao2017fashion}. CIFAR-10 consists of RGB images of size $32\times32$ with 50,000 training and 10,000 test images, while FashionMNIST contains grayscale images of $24\times24$ with 60,000 training and 10,000 test images. Two experimental settings are used to evaluate the algorithms. {\em In Setting 1}, the first four classes from each dataset are selected with sample sizes of 250, 5,000, 250, and 250, respectively. {\em In Setting 2}, the first nine classes each have 250 samples, while the tenth class has 5,000 samples. These settings are designed to simulate different class distributions and assess algorithm performance under varying levels of class imbalance, reflecting real-world scenarios where such imbalances are common.

{\bf Training Model.} We employ the ResNet-18 architecture for our experiments, implemented using the PyTorch framework. The ResNet-18 model consists of four residual blocks, each containing two convolutional layers followed by Batch Normalization layers. Specifically, each block includes a sequence of four layers: Conv2d, BatchNorm2d, ReLU, and Conv2d, repeated consistently across all layers. After these blocks, a Fully Connected layer is applied at the end of the network to perform classification. This structure ensures that the model benefits from deep residual learning while maintaining a manageable level of complexity. 

We use the Adam optimizer with a learning rate of 0.001. Additionally, we apply image preprocessing techniques such as \textit{RandomCrop}, \textit{HorizontalFlip}, and \textit{Normalization} to the raw images.


{\bf Baselines.} We consider four baselines to compare with our proposed approach, FisherMask, as described below:

\begin{itemize}
    \item \textbf{Random Sampling \cite{ash2019deep}.} Certain points are chosen in a naive manner and added to the unlabeled dataset. 
    \item \textbf{Entropy Sampling \cite{safaei2024entropic}.} A traditional AL approach that selects unlabeled instances with the highest entropy. 
    \item \textbf{BAIT Sampling \cite{ash2021gone}.}  Fisher-based active selection method that selects batches of samples by optimizing a bound on the MLE (Maximum Likelihood Estimation) error in terms of the last layer of Fisher information matrices. 
    \item  \textbf{Margin Sampling \cite{wang2014new}.} A technique used to select samples based on the minimal difference between the top two predictions for each class. 
    \item  \textbf{K-center Greedy \cite{settles2009active}.} An approach that chooses $k$ samples by solving a k-center problem on $z_{x}$ where $z_{x}$ is the embedding of $x$ derived from the penultimate layer.
\end{itemize}

\begin{figure}
    \centering
    \begin{subfigure}[b]{0.45\textwidth}
        \centering
        \includegraphics[width=0.9\textwidth]{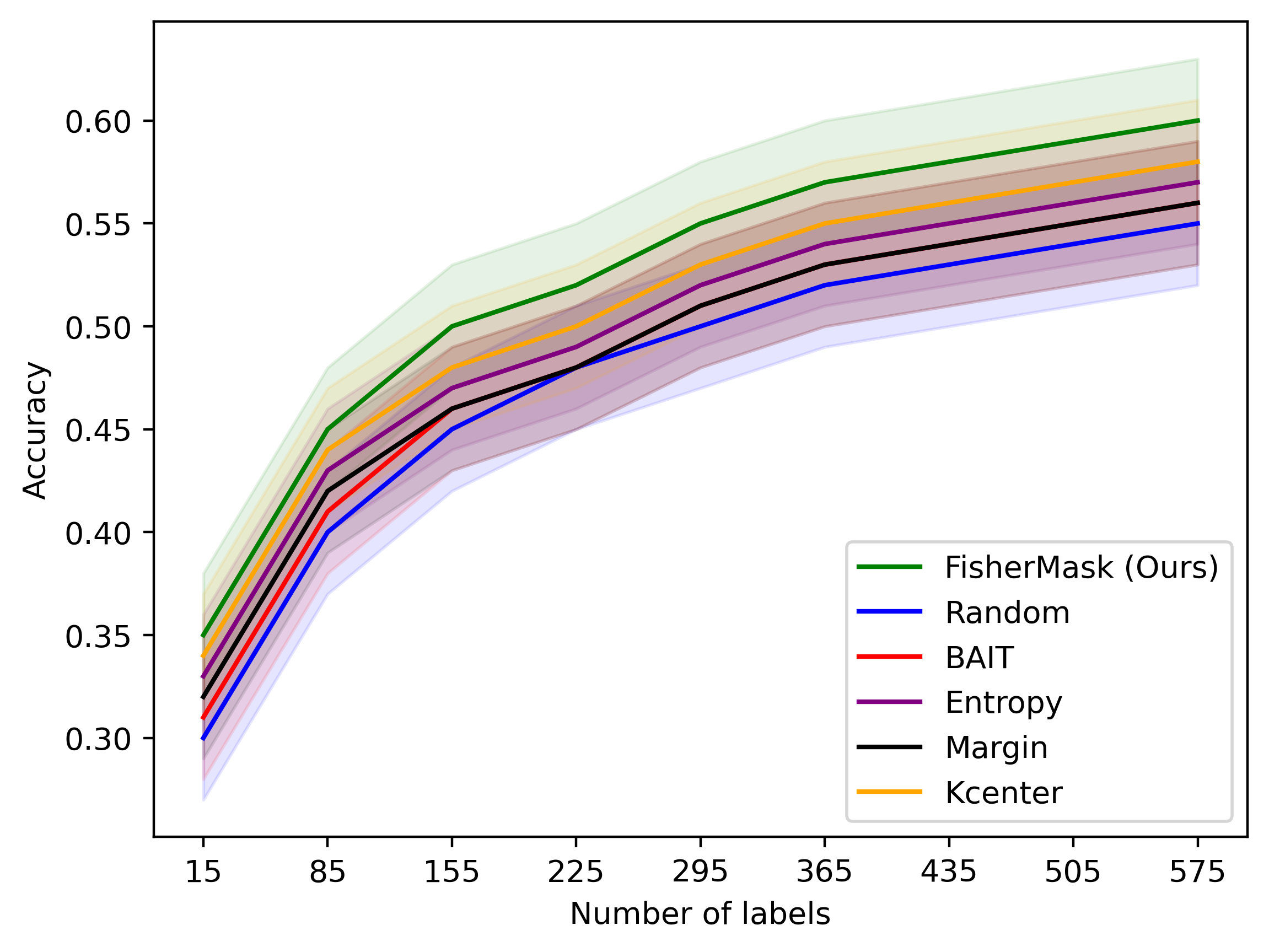}
        \caption{Low.}\label{fig:4}
    \end{subfigure}
    \begin{subfigure}[b]{0.45\textwidth}
        \centering
        \includegraphics[width=0.9\textwidth]{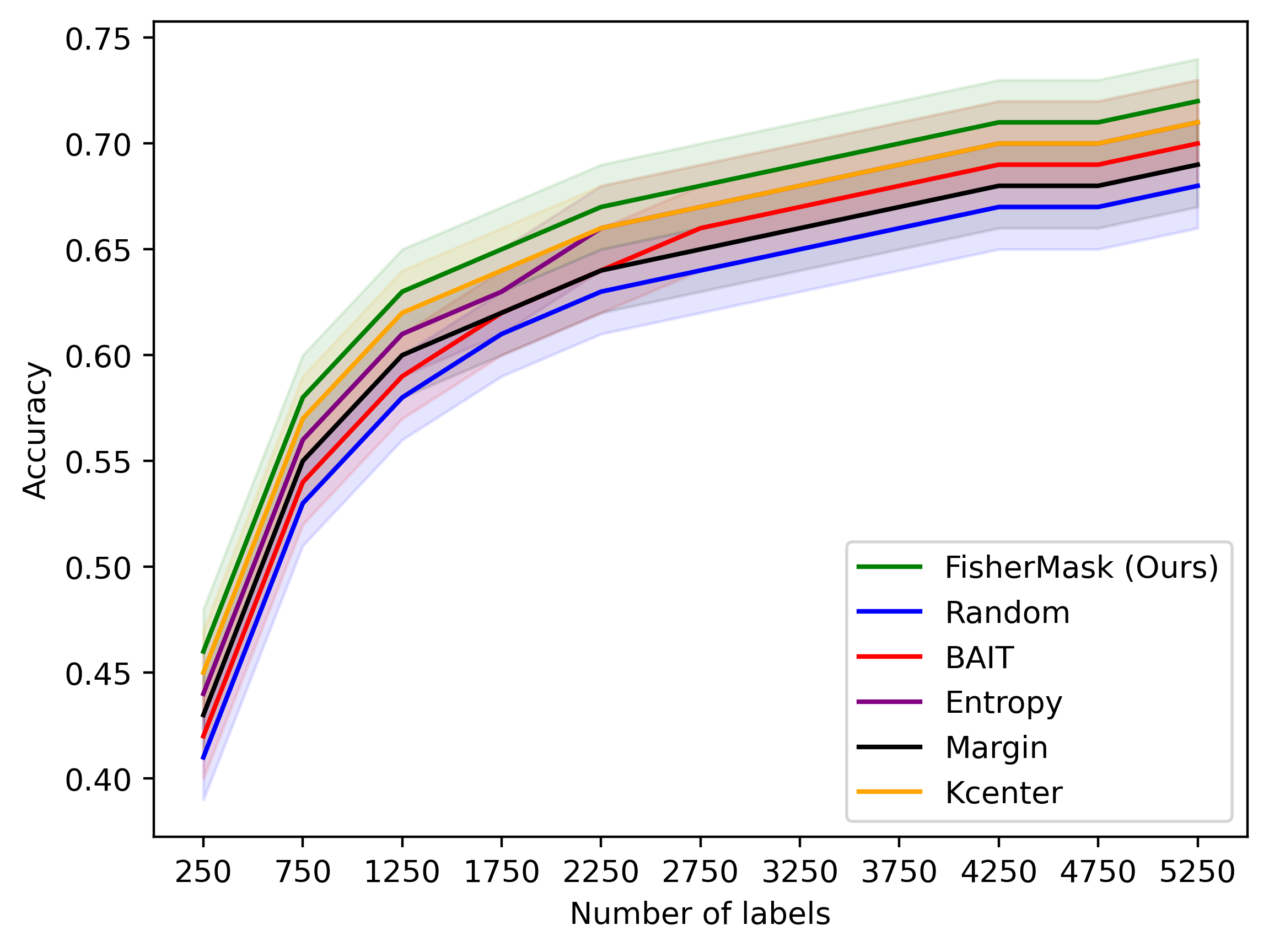}
        \caption{High.}\label{fig:5}
    \end{subfigure}
    \caption{Data regime for imbalanced CIFAR10.}  \label{fig:cifareim}
\end{figure}
{\bf Experimental Results.}

{\bf 1) Setting \#1.} In a scenario with limited/low data availability, as depicted in \fref{fig:4}, we evaluate various approaches using a subset of 575 points from a pool of 5750 samples to address two primary challenges. The first challenge involves an imbalanced dataset, characterized by uneven distribution of samples across different classes. The second challenge relates to the limited sample size, where only 10\% of the entire imbalanced set is utilized. While entropy generally remains below BAIT for a significant portion of the graph, it shows a slight increase towards the end of the AL rounds. The Random Sampling approach consistently performs poorly due to its naive approach of randomly selecting samples. The k-center Greedy approach initially performs similarly to other strategies but steadily improves throughout the graph. Margin Sampling initially performs worse than Random Sampling but gradually improves. Notably, {\em FisherMask consistently outperforms baselines across a substantial part of the plotted data, highlighting its superior performance}\footnote{The shaded regions on the learning curves represent the variance, providing an indication of the uncertainty or spread in the performance measurements.}.

In \fref{fig:5}, we observe the performance in a high-data regime, where the entire unlabeled dataset is utilized by the end of the AL rounds. The process starts with 250 randomly chosen samples and incrementally adds 250 more until all 5,750 samples from the imbalanced CIFAR-10 dataset are used. The graph shows that all algorithms begin with an average accuracy of approximately 45\% and improve to about 70\% by the end of the cycles. FisherMask consistently outperforms margin sampling, k-center greedy, and BAIT. Entropy sampling, however, maintains performance comparable to FisherMask throughout the AL process. Random sampling shows initially lower performance but steadily improves, eventually surpassing its initial accuracy by the end of the cycles.

We also examined both the low and high data regimes on the FashionMNIST dataset, which exhibited similar trends to those observed with the CIFAR-10 dataset.


{\bf 2) Setting \#2.} In this scenario, samples are selected from a different set of classes. Specifically, 250 samples are chosen from each of the first nine classes, while the tenth class contains 5,000 samples for both datasets. This setup simulates real-world situations where one class significantly outweighs the others, leading to potential bias in an ML model towards the predominant class and resulting in skewed decision-making due to its overrepresentation.

\begin{figure}
    \centering
    \begin{subfigure}[b]{0.45\textwidth}
        \centering
        \includegraphics[width=0.9\textwidth]{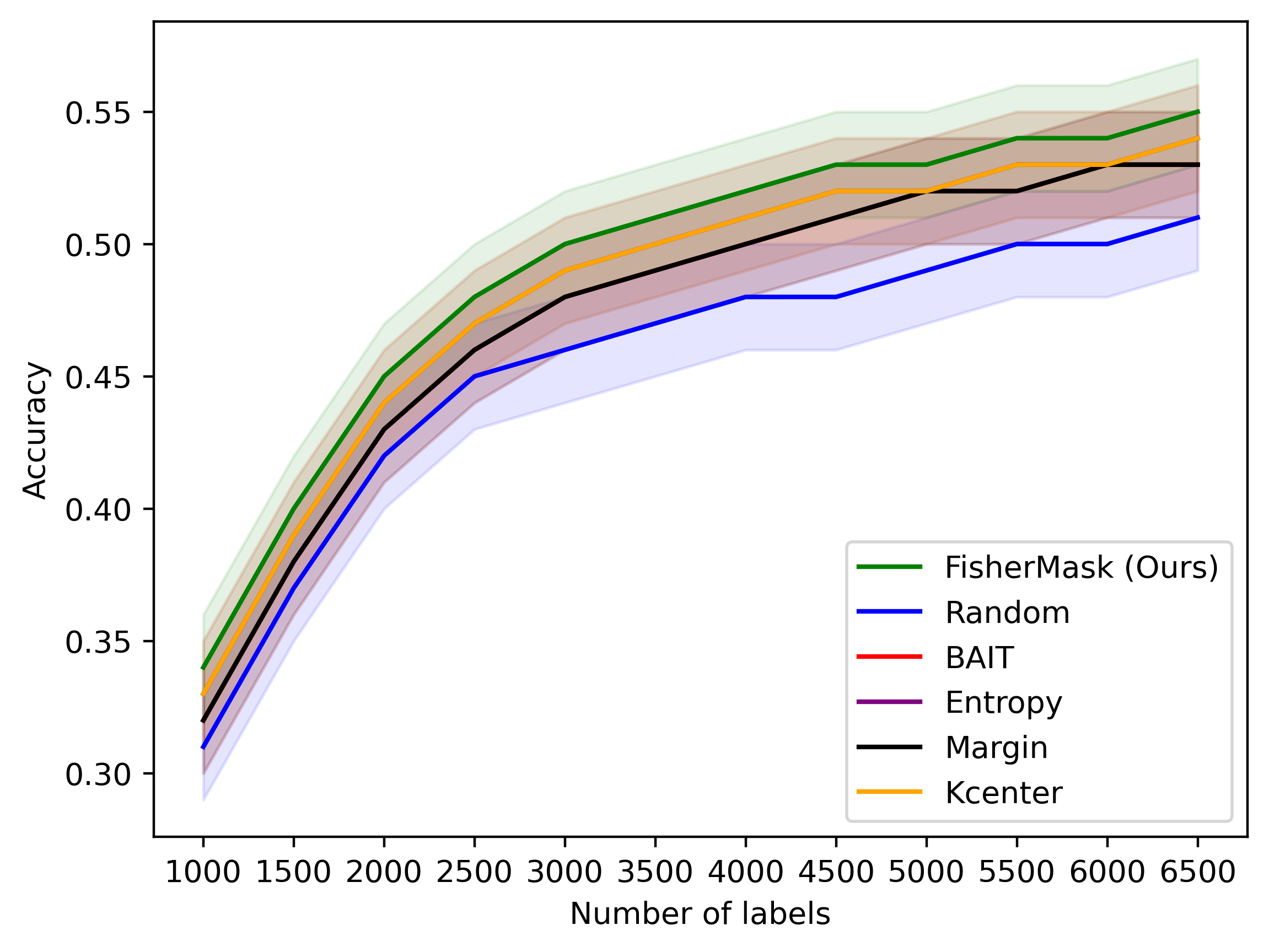}
        \caption{CIFAR10.}\label{fig:CIF_im}
    \end{subfigure}
    \begin{subfigure}[b]{0.45\textwidth}
        \centering
        \includegraphics[width=0.9\textwidth]{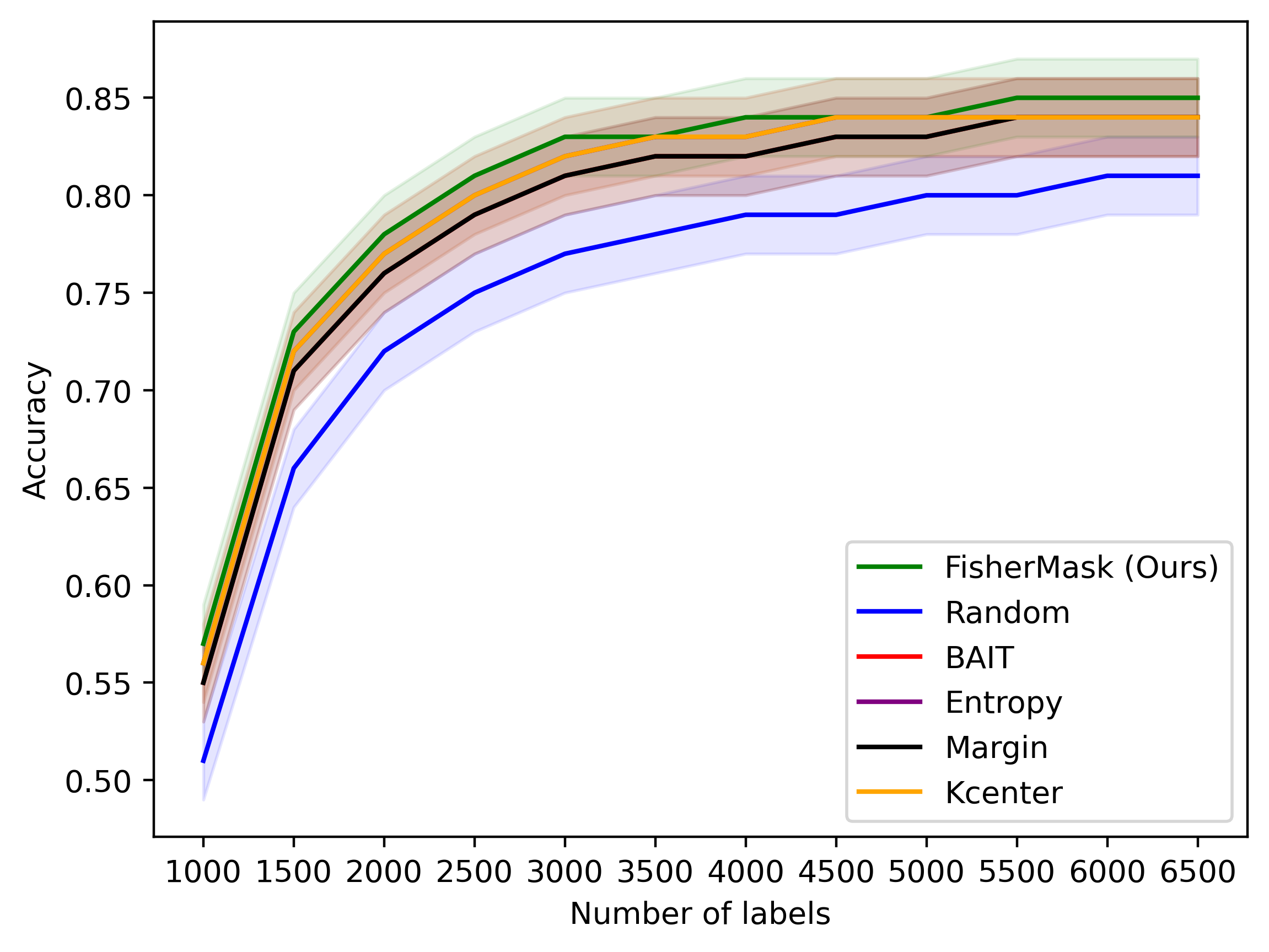}
        \caption{FashionMNIST.}\label{fig:FMNIST_im}
    \end{subfigure}
    \caption{Result for Imbalanced datasets.}  \label{fig:GMM_Sensor}
\end{figure}

\fref{fig:CIF_im} displays the average accuracy curves for the imbalanced CIFAR-10 dataset under this setting. A batch size of 500 is used until the budget of 6,500 is exhausted. All algorithms start with an accuracy of approximately 30\% and achieve around 60\% mean accuracy by the end of the AL cycles. FisherMask consistently outperforms the other methods throughout the experiment. BAIT shows a notable improvement around the midpoint but underperforms in the latter stages. Entropy and Margin sampling exhibit alternating performance relative to each other. K-center Greedy initially lags but ultimately surpasses Random sampling and BAIT by the end of the cycles. Random sampling remains the least effective strategy throughout the experiment.

Similarly, in \fref{fig:FMNIST_im}, we present the results for the FashionMNIST dataset under this imbalanced data setting. As with the previous scenario, 500 samples are added to the training data in each AL round until the budget of 6,500 is reached. All techniques start with an average accuracy of approximately 52\% and achieve around 83\% accuracy by the end of the cycles. The increase in accuracy across all strategies can be attributed to FashionMNIST’s grayscale images, which are generally easier for models to learn. Consequently, the differences in learning curves among the strategies are minimal.



\section{Conclusion and Future work}

In this paper, we proposed FisherMask, a novel technique that uses a sparse mask of weights to identify the most impactful samples based on their Fisher Information values. By selecting the top parameters across the entire network, FisherMask determines which samples to update and use for the next batch in an AL round. Experimental results show that FisherMask performs well under label sparsity and challenging class imbalances. Both experimental and theoretical analyses demonstrate that our approach outperforms existing baselines, particularly in low-data regimes. For future work, we aim to explore the effectiveness of FisherMask on additional datasets that closely mimic real-world scenarios.

In our future work, we plan to leverage the Fisher Information Matrix (FIM) to enhance sample selection, as FIM captures critical information about model parameters, leading to more informed choices. By selecting samples that maximize the expected information gain about these parameters, FIM can help identify those samples that provide the most insight into parameter estimation. This approach aligns with optimal experimental design criteria: for instance, A-optimality minimizes the average variance of parameter estimates, while D-optimality maximizes the determinant of the FIM, thereby improving the precision of estimated parameters. Additionally, we are considering integrating FIM into the loss function used during AL, allowing us to reward or penalize the selection of samples based on their informativeness.

\bibliographystyle{IEEEtran}

\bibliography{IEEEabrv,./references.bib}



\end{document}